\documentclass[authoryear]{elsarticle}

\usepackage{subcaption} 
\usepackage{amssymb}
\usepackage{tabularx}
\usepackage{tikz, pgfplots}
\usepackage{booktabs}
\usepackage{amsmath, amssymb}
\usepackage[table]{xcolor}  
\definecolor{best}{HTML}{ADD8E6}    
\definecolor{second}{HTML}{FFFACD}  
\usepackage{amsmath}
\usepackage[authoryear]{natbib}
\setcitestyle{round, citesep={,}} 
\usepackage{multirow}

\journal{IMAVIS}

\begin{document}

\begin{frontmatter}



\title{OSCAR: Open-Set CAD Retrieval from a Language Prompt and a Single Image} 


\author[label1]{Tessa Pulli}
\author[label2]{Jean-Baptiste Weibel}
\author[label1]{Peter Hönig}
\author[label1]{Matthias Hirschmanner}
\author[label1]{Markus~Vincze}
\author[label2,label3]{Andreas Holzinger}

\address[label1]{Automation and Control Institute, TU Wien, Wien, Austria\\
\texttt{\{last\_name\}@acin.tuwien.ac.at}}
\address[label2]{BOKU University, Human-Centered AI Lab, FTEC, Department for Ecosystem Management, Climate and Biodiversity, Wien, Austria\\
\texttt{\{first\_name\}.\{last\_name\}@boku.ac.at}}
\address[label3]{Institute for Human Centered Computing, Faculty of Informatics and Biomedical Engineering, TU Graz, Graz, Austria\\
\texttt{a.holzinger@tugraz.at}}
\begin{abstract}
6D object pose estimation plays a crucial role in scene understanding for applications such as robotics and augmented reality.
To support the needs of ever-changing object sets in such context, modern zero-shot object pose estimators were developed to not require object-specific training but only rely on CAD models.
Such models are hard to obtain once deployed, and a continuously changing and growing set of objects makes it harder to reliably identify the instance model of interest.
To address this challenge, we introduce an \textbf{O}pen-\textbf{S}et \textbf{CA}D \textbf{R}etrieval from a Language
Prompt and a Single Image (OSCAR), a novel training-free method that retrieves a matching object model from an unlabeled 3D object database.
During onboarding, OSCAR generates multi-view renderings of database models and annotates them with descriptive captions using an image captioning model.
At inference, GroundedSAM detects the queried object in the input image, and multi-modal embeddings are computed for both the Region-of-Interest and the database captions. OSCAR employs a two-stage retrieval: text-based filtering using CLIP identifies candidate models, followed by image-based refinement using DINOv2 to select the most visually similar object. 
In our experiments we demonstrate that OSCAR outperforms all state-of-the-art methods on the cross-domain 3D model retrieval benchmark MI3DOR.
Furthermore, we demonstrate OSCAR's direct applicability in automating object model sourcing for 6D object pose estimation.
We propose using the most similar object model for pose estimation if the exact instance is not available and show that OSCAR achieves an average precision of 90.48\%  during object retrieval on the YCB-V object dataset. 
Moreover, we demonstrate that the most similar object model can be utilized for pose estimation using Megapose achieving better results than a reconstruction-based approach.
\end{abstract}
\begin{keyword}
3D Model Retrieval \sep 6D Object Pose Estimation \sep Image-based Object Retrieval


\end{keyword}



\end{frontmatter}



\section{Introduction}
\label{sec:intro}
\noindent{}6D object pose estimation is a key task in robotics, and Augmented Reality.
Instance-level approaches~\citep{wang2021gdr-net} are commonly used and have shown impressive results on known objects. They however, require a re-training to handle any additional unseen instances.
Recently, zero-shot pose estimation methods emerged to estimate 6D object poses of novel instances~\citep{labbe2022megapose, ausserlechner2024zs6d, foundationposewen2024, lin2024sam6d, thalhammer2023self}.
Existing zero-shot methods can be divided into model-based~\citep{labbe2022megapose, ausserlechner2024zs6d, foundationposewen2024, lin2024sam6d} and model-free approaches~\citep{foundationposewen2024, Shugurov_2022_CVPR, caraffa2024freeze, lee2024mfos}, depending on whether a 3D mesh is required.
Model-based methods typically yield better performance results~\citep{hodan2024bop} but assume the availability of a ground-truth 3D object model.
Acquiring these CAD models is a time-consuming process that requires special equipment or skilled personnel, and considerable pre-processing before they can be included as a reference model, limiting the feasibility of modeling once the autonomous agent is deployed.
\noindent{}Furthermore, maintaining and utilizing a large, dynamic database of 3D object models presents significant challenges for open-set 6D pose estimation. 
Even when models are available, they are often unlabeled or categorized only by coarse classes. 
The manual effort required to accurately label and maintain thousands of object models in large databases like in Megapose~\citep{labbe2022megapose}, is prone to error.

\noindent{}While CAD model retrieval is an established research direction~\citep{gumeli2022roca, gao2024diffcad}, 3D object model retrieval for 6D object pose estimation is largely unexplored. 
To address this problem, we propose a novel method that retrieves a corresponding 3D object model from an unlabeled database using an image and text input. 
\begin{figure}[tbp]
  \centering
  \includegraphics[width=0.8\columnwidth]{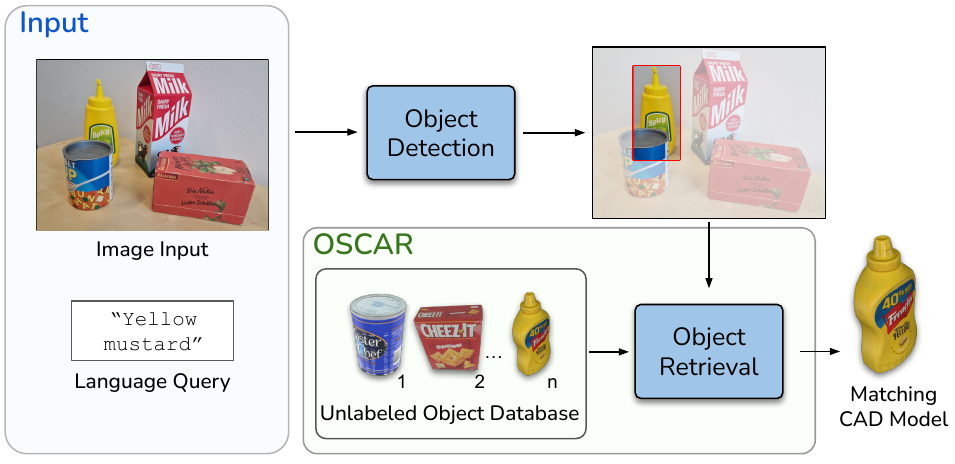}
  \caption{Overview: We propose \textbf{OSCAR} to retrieve a 3D object model from an unlabeled database of 3D objects with a single image input and a language prompt.}
  \label{fig:overview}
\end{figure}
In this paper, we introduce \textbf{OSCAR}, short for \textbf{O}pen-\textbf{S}et \textbf{CA}D \textbf{R}etrieval from a Language Prompt and a Single Image (see Fig. \ref{fig:overview}). 
During the onboarding stage, each 3D model of the database is rendered from multiple viewpoints, and an image captioning model generates descriptive texts for each view. 
These textual descriptions are stored alongside the models to enable language-guided retrieval without manual annotation.
During inference, we segment the Region-of-Interest (ROI) from an RGB image using GroundedSAM~\citep{ren2024grounded} with a language prompt. 
OSCAR relies on a two-stage object retrieval:
Firstly, we compute the textual embeddings for each description of the rendered views, together with the image embedding of the RoI using CLIP.
Candidates whose cosine similarity passes a defined threshold are selected.
For each candidate, the ROI embedding is compared to embeddings of pre-rendered views of the CAD model using DINOv2.
With this strategy, we select the model with the highest similarity, considering textual embeddings and image embeddings.
In summary, our contributions are the following:

\begin{itemize}
    \item We demonstrate that the fusion of vision-language embeddings (CLIP) and self-supervised visual embeddings (DINOv2) is superior to geometry or purely visual feature-based retrieval methods, achieving state-of-the-art performance on the cross-domain 3D model retrieval benchmark MI3DOR.
    \item We show, that utilizing a clean, retrieved 3D model yields superior 6D pose estimation accuracy and better efficiency compared to approaches relying on on-the-fly reconstructed object models.
    demonstrate the effective integration of OSCAR with MegaPose, achieving superior 6D pose estimation accuracy over reconstruction baselines, thereby proving the importance of high-quality models in object pose estimation.
  \item We embed these findings in a novel training-free framework, OSCAR, combining multi-modal 3D object model retrieval from an unlabeled 3D object database using an RGB image and a language prompt.

\end{itemize}

\noindent{}We evaluate OSCAR on the cross-domain 3D model retrieval benchmark MI3DOR~\citep{mi3dor} and show that we outperform existing methods, e.g.~\citep{dlea, sc-ifa, s2mix}.
The paper proceeds as follows:  We first review the related work in 6D object pose estimation, 3D reconstruction, and 3D model instance retrieval in Section II. In Section II, we detail our novel OSCAR framework, including the multi-modal onboarding stage and the two-stage retrieval process. Section IV presents the experimental setup, covering the datasets and evaluation metrics used. We then present our state-of-the-art results on the MI3DOR benchmark and our integration with MegaPose in Section V. Finally, we conclude the paper in Section VI.

\section{Related Work}
\label{sec:sota}
\noindent{}In this section, we discuss the state-of-the-art in three key areas relevant to this work: 6D object pose estimation, 3D reconstruction, and 3D model instance retrieval.
\subsection{6D Object Pose Estimation}
\label{sec:sota_pose_estimation}
\noindent{}6D Object Pose Estimation plays a crucial role in robotics and Augmented Reality, enabling precise object manipulation and interaction.
Instance-level methods~\citep{wang2021gdr-net} achieve strong performance on objects encountered during training but struggle to generalize to novel instances.
Zero-shot pose estimation approaches have been developed to address this challenge by estimating 6D poses of novel object instances~\citep{labbe2022megapose, ausserlechner2024zs6d, foundationposewen2024, lin2024sam6d}.
The BOP challenge~\citep{hodan2024bop} categorizes 6D object pose estimation into model-based~\citep{labbe2022megapose, ausserlechner2024zs6d, foundationposewen2024, lin2024sam6d} and model-free approaches~\citep{foundationposewen2024, Shugurov_2022_CVPR, caraffa2024freeze, lee2024mfos, 6dof}, depending on whether a 3D mesh is required.
Model-based methods typically yield better performance but rely on the availability of a ground-truth 3D object model~\citep{burde2025wacv}.
Acquiring these CAD models is a time-consuming process that requires special equipment, skilled personnel, and considerable pre-processing before it can be integrated into a 6D object pose estimation method.
They assume that database models are pre-labeled and demand manual effort to manage an object database.
In a practical setting, such as household environments, it is infeasible to acquire CAD models for each object.
A promising alternative is model-free approaches, which do not require a mesh at inference but reconstruct these models on the fly through reference images~\citep{foundationposewen2024} or videos~\citep{sun2022onepose}.
\subsection{3D Reconstruction}
\label{sec:reconstruction}
\noindent{}Several approaches combine object pose estimators with 3D reconstruction methods.
Traditional multi-view stereo and structure-from-motion methods~\citep{wang2024vggsfm} provide accurate results but rely on dense observations from multiple viewpoints, which often require controlled settings and are computationally expensive.
Recent learning-based methods employ neural implicit representations to reconstruct 3D shapes from multiple views~\citep{mildenhall2021nerf, APS-NeuS}. 
Despite improved visual quality, these approaches typically require a large number of input images and runtimes up to several hours, limiting their applicability in real-time scenarios.
Diffusion-based NVS~\citep{xiang2025structured, long2024wonder3d} methods overcome this limitation as they require only a few images to reconstruct a watertight mesh.
While promising in terms of speed, these methods often synthesize unseen parts of the object and do not preserve true scale or geometry.
Given these limitations, we propose an alternative approach to retrieve the most similar object from a pre-existing 3D model database, rather than performing on-the-fly reconstruction.
\subsection{3D Model Instance Retrieval}
\noindent{}3D model retrieval and alignment is a relevant problem in several domains and aims to find the most similar 3D model in a database given a query.
Traditional approaches can be broadly categorized into model-based, view-based, and feature-fusion methods.
Model-based methods directly operate on 3D data such as meshes, point clouds~\citep{qi2017pointnet, 3d_shape_retrieval}, or voxels~\citep{wang2019normalnet}, learning geometric features that encode spatial and structural information.
These methods, for instance, are applied in indoor scene acquisition~\citep{gumeli2022roca, gao2024diffcad} to replace noisy 3D meshes with clean CAD models.
While these approaches perform well with point clouds, they are not directly applicable to RGB-only scenes.

\noindent{}View-based methods address this limitation by rendering multiple 2D views of a 3D model and applying image-based feature extraction. 
MVCNN~\citep{su2015mvcnn} employs 2D CNNs to process each rendered view and aggregates them through view pooling into a compact 3D descriptor.
GVCNN~\citep{feng2018gvcnn} further improves on this by grouping view features and hierarchically pooling them to obtain richer global representations.
Later, MVTN~\citep{hamdi2021mvtn} leverages differentiable rendering to learn optimal viewpoints for 3D shape recognition.
These approaches effectively bridge the gap between 2D and 3D domains, but they still require synthetic rendering pipelines or pre-existing 3D models, which can be impractical for real-world scenarios.

\noindent{}Beyond geometry-based methods, recent work explores cross-modal retrieval, where the query comes from a different modality such as images or sketches~\citep{pagml}.
Sketch-based retrieval methods like PAGML~\citep{pagml} employ precise alignment-guided metric learning to reduce discrepancies between sketch and 3D shape domains.
Similarly, CNOS~\citep{nguyen2023cnos} reformulates instance recognition as a retrieval problem, comparing visual embeddings from RGB images against a database of known objects.
While effective for category-level retrieval, such purely visual methods struggle in open-set conditions, where unseen objects must be recognized.

\noindent{}To address these challenges, advances in multi-modal representation learning have introduced joint image–text embeddings that align visual and linguistic information in a shared semantic space.
Models such as CLIP~\citep{radford2021clip}, BLIP~\citep{li2022blip}, and DINO~\citep{oquab2024dinov2} learn cross-modal representations that generalize well across domains and modalities.
These embeddings have proven particularly valuable in robotic and household settings, where visual tasks, such as identifying or retrieving objects are often guided by natural-language instructions~\citep{Freiring:2025:HumanRobot} and have also been used in object retrieval tasks~\citep{song2025adaptive}. 

\noindent{}This progress has led to dedicated benchmarks like MI3DOR~\citep{mi3dor} (Multi-modal Image-to-3D Object Retrieval), which specifically tests the generalization capacity of retrieval methods in open-set, cross-domain scenarios. The current state-of-the-art on MI3DOR is largely dominated by methods like S2Mix~\citep{s2mix}, SC-IFA~\cite{sc-ifa}, and DLEA~\cite{dlea}, which primarily employ domain adaptation or feature fusion techniques to handle the large domain gap between 2D query images and 3D model views. Despite their efforts, these methods often rely on complex training procedures or fail to fully leverage the dense semantic information available in multi-view renderings.

\noindent{}Building on these insights, we investigate the use of text, image, and joint multimodal embeddings to retrieve corresponding 3D models or instances from an object database.


\section{Method}
\label{sec:method}
\begin{figure*}[h]
\begin{center}
  \includegraphics[width=\textwidth]{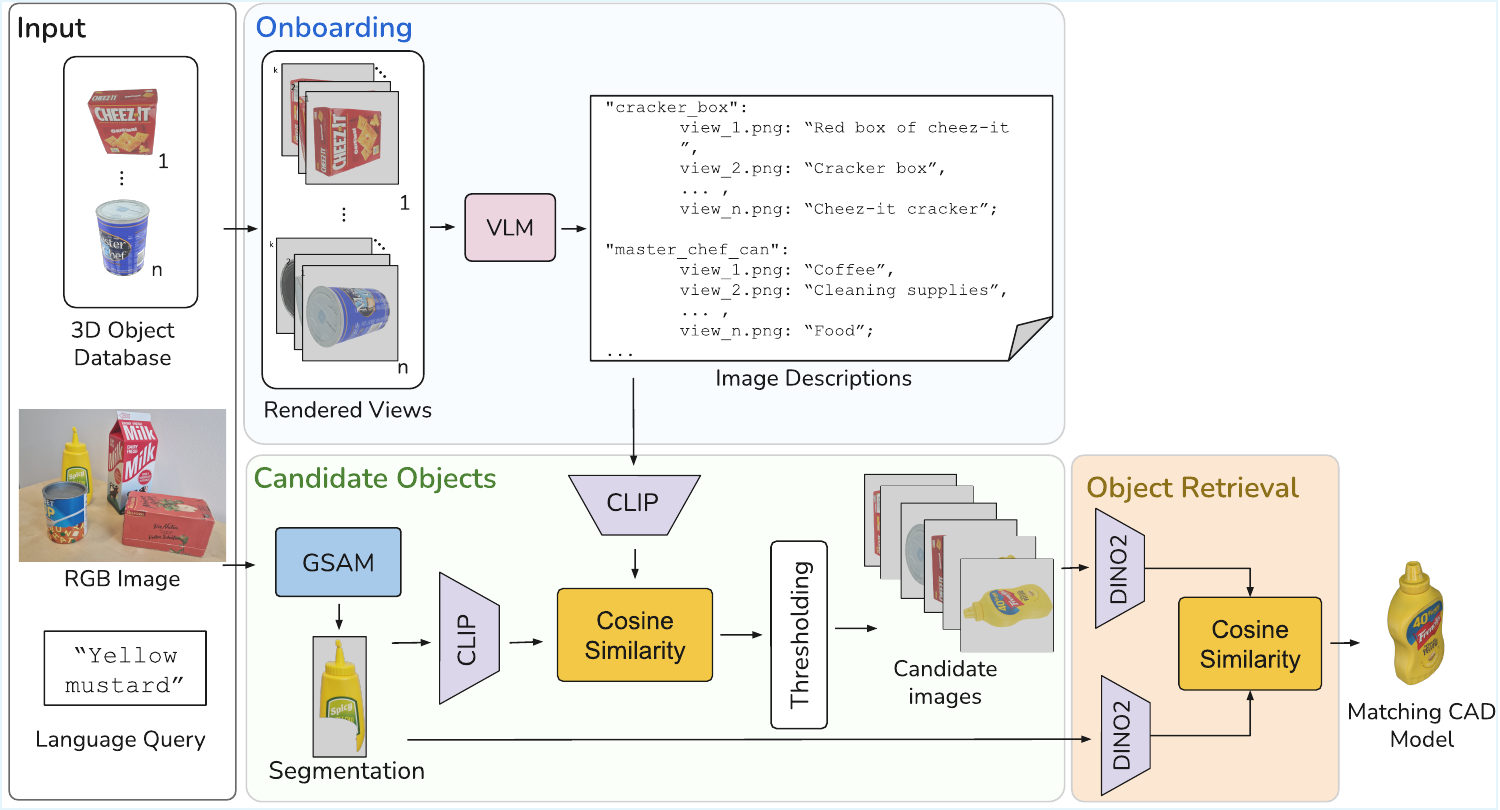}
  \caption{\textbf{Concept overview:} During the onboarding stage, each 3D model is rendered from multiple views and automatically captioned to enable language-guided retrieval without manual labels. During the inference stage, a Region-of-Interest (ROI) is generated from an RGB image and a language prompt using GroundedSAM. First, CLIP embeddings of the ROI and model captions are compared via cosine similarity and filtered according to a threshold. The DINOv2 embeddings are computed for these candidate images. Based on the cosine similarity, the object model with the highest similarity is selected.}
  \label{fig:method}
\end{center}
\end{figure*}

\noindent{}With an input RGB image \( I \in \mathbb{R}^{H \times W \times 3} \), a language prompt  \(L\), and 
a database of 3D CAD models \( \mathcal{S} = \{s_1, s_2, \dots, s_N\} \), we aim to retrieve the CAD model \( s^*\)  with the highest similarity to the query object. 
An overview of the approach is given in Fig.~\ref{fig:method}.

\subsection{Onboarding Stage}
\noindent{}Following the approach proposed by CAP3D~\citep{luo2023_scalable}, we render for each CAD model of the database \( s_i \in \mathcal{S} \) synthetic images \(
\mathcal{R}_i = \{ r_{i1}, r_{i2}, \dots, r_{iK} \}
\) from \( K = 8 \) viewpoints, with two views from an elevation angle \( K_{1,2,3,4} = -15^\circ\) and \( K_{5,6,7,8} = +15^\circ \), while the azimuth angles are distributed evenly around the object.
Each rendered image \( r_{ij} \) is captioned using LLaVA-v1.5-7B~\citep{liu2023improvedllava}, producing 
\(\mathcal{C}_i = \{ c_{i1}, c_{i2}, \dots, c_{iK} \}\) descriptions. \\
When a novel object is added to the database, OSCAR verifies whether the corresponding CAD model is already associated with a rendered view and a textual description. 
If such data is missing, the onboarding pipeline automatically renders the required viewpoints and generates the associated captions. 
\subsection{Object Retrieval}
\noindent{}We assume a given input RGB image 
\( I \in \mathbb{R}^{H \times W \times 3} \) and a natural language prompt \( L \in \mathcal{L} \). 
During the detection stage, we utilize GroundedSAM~\citep{ren2024grounded} to generate a bounding box around the RoI and segment the corresponding object and a gray background to match the rendered images.
The object retrieval stage consists of two phases: (1) filtering of candidate objects based on descriptions and query embedding, and (2) the retrieval of the most similar object based on DINOv2~\citep{oquab2024dinov2} embeddings. 
The objective of the first stage is to filter candidate objects at the textual level. 
In our experiments, we find that image embeddings are particularly effective for datasets containing distinct objects, as they capture fine-grained texture similarities.
However, when objects share similar textures, image embeddings may match entirely different categories.
To mitigate this issue, we propose a sanity check in which CLIP text embeddings are first used to filter plausible candidates based on semantic similarity. 
Only then are image embeddings applied to refine the selection, ensuring that the final match is both semantically and visually consistent.
\paragraph{\textbf{Filtering of Candidates}}
\noindent{}At inference time, textual descriptions \(c_{ij}\) are encoded into embeddings using a CLIP text encoder \( f_L \), giving text embeddings

\[ \mathbf{t}_{ij} = f_L(c_{ij}) \in \mathbb{R}^d \] 

\noindent{}At the same time, we encode the ROI using CLIP as an image encoder \( f_I \), producing an image embedding 
\[ \mathbf{q_{CLIP}} = f_I(I_{\text{ROI}}) \in \mathbb{R}^d \]
We compute the cross-modal cosine similarity between the image embedding \(\mathbf{q}\) and each caption embedding \(\mathbf{t}_{ij}\). For each object \(s_i\), we aggregate over its captions:
\[
\mathrm{sim}_{\text{text}}(s_i) \;=\; \max_{j}\; \frac{\langle \mathbf{q_{CLIP}}, \mathbf{t}_{ij} \rangle}{\|\mathbf{q_{CLIP}}\| \,\|\mathbf{t}_{ij}\|}.
\]
We then form a candidate set \(\mathcal{S}'\) by selecting all objects with \(\mathrm{sim}_{\text{text}}(s_i) \ge \tau_{\text{text}}\) (threshold \(\tau_{\text{text}}\) with \(\tau_{\text{text}} = 0.37\).
If \(\mathcal{S}'\) is empty (no candidate passes \(\tau_{\text{text}}\)), we fall back to the top-\(k\) text candidates.

\paragraph{\textbf{Image-based Retrieval}}
\noindent{}For each candidate \(s_i \in \mathcal{S}'\) we compare the ROI against pre-rendered views \(\{r_{ik}\}\) of the corresponding CAD model. Each rendered view is embedded by the DINOv2 image encoder:
\[
\mathbf{v}_{ik} = f_I(r_{ik}) \in \mathbb{R}^d.
\]
We score a candidate by the best image–image similarity across its views:
\[
\mathrm{sim}_{\text{img}}(s_i) \;=\; \max_{k}\; \frac{\langle \mathbf{q_{DINO}}, \mathbf{v}_{ik} \rangle}{\|\mathbf{q_{DINO}}\| \,\|\mathbf{v}_{ik}\|}.
\]

The final retrieved image (and thus the CAD model) is chosen as

\[
c^{\ast} \;=\; \arg\max_{s_i \in \mathcal{S}'} \mathrm{sim}_{\text{img}}(s_i),
\]

and \(c^{\ast}\) is mapped to its corresponding CAD model.

\section{Experiments}
\label{sec:exp}
\noindent{}We evaluate our approach and its components against state-of-the-art baselines on image-based 3D object retrieval and 6D pose estimation.
The following experiments were conducted on a system equipped with an NVIDIA GeForce RTX 4090 GPU and an AMD Ryzen 9 5900X CPU.
\subsection{Datasets}
\noindent{} We assess OSCAR on the 3D model datasets (MI3DOR~\citep{mi3dor}), and on three datasets for object retrieval in the context of 6D object pose estimation (YCB-V~\citep{xiang2018posecnn}, Housecat6D~\citep{jung2024housecat6d}, and YCB-V combined with GSO~\citep{downs2022google}).
MI3DOR is designed for monocular image-based 3D model retrieval and includes 2D images and 3D models across 21 categories. 
The 2D images, serving as queries, are adapted from ImageNet.

\noindent{}To show OSCAR's capabilities for 6D object pose estimation in domestic environments, we consider the following datasets: YCB-V, consisting of 21 household objects, and Housecat6D, containing 194 diverse objects across 10 household categories.
Additionally, we assess OSCAR’s scalability and robustness by using the Google Scanned Objects (GSO) dataset as distractor objects.
We combine the YCB-V dataset with the GSO objects, which are 1030 CAD models of common household objects.

\subsection{Evaluation Metrics}
\noindent{}In the following, we introduce the metrics to evaluate the object retrieval capabilities and the application in object pose estimation.

\subsection{Mean Average Precision (mAP@k)}
\noindent{}To evaluate our object retrieval strategy, we adopt the mean Average Precision at top-$k$ (mAP@$k$).
For each query, the retrieved results are ranked by their cosine similarity score.
The Average Precision (AP) for a query is defined as:
\[
mAP@K = \frac{1}{n} \sum_{i=1}^{n} AP@k,
\]
where $n$ denotes the total number of queries, and $AP@k$ represents the average precision of the top-$K$ ranked items for the query $u$. 
This metric accounts for both the relevance and the ordering of recommendations, as higher-ranked relevant items contribute more significantly to the score. 
\subsection{Earth Mover’s Distance}
The general definition of the Earth Mover’s Distance (EMD) between two discrete distributions is
\[
\mathrm{EMD}(\alpha, \beta) = \min_{\pi \in \Pi(\alpha, \beta)} \sum_{i=1}^{N} \sum_{j=1}^{M} \pi_{i,j} \, c(x_i, y_j),
\]
where \(\pi \in \mathbb{R}^{N \times M}\) is a non-negative matrix representing the optimal correspondence between points \(x_i \in X\) and \(y_j \in Y\), and \(c(\cdot,\cdot)\) is a distance function, typically chosen as \(c(x_i, y_j) = \frac{1}{2}\|x_i - y_j\|_2^2\). 
Intuitively, this represents the weighted distance between all pairs of points according to the optimal correspondence \(\pi\).  
\subsection{Translation Error}
\noindent{}The translation error is defined as the Euclidean distance between the predicted translation vector 
$\hat{t}$ and the ground truth translation $t$:
\[
e_t = \lVert \frac{1}{n} \sum_{i=1}^{n=3} (t_{1,i} - t_{2,i}) \rVert
\]
\subsection{Rotation Error}
The rotation error is computed as the angular difference between the predicted rotation matrix 
$\hat{R}$ and the ground truth $R$:
\[
e_R = \lVert R_1 \, R_2^\top \rVert_2^2
\]

\section{Results}
\noindent{}The following section presents the main results and ablation studies.
\subsection{Object Retrieval}
\noindent{}Table~\ref{tab:main-results} reports the results of several object retrieval methods on the MI3DOR dataset. 
We employ six commonly used evaluation criteria to
comprehensively assess retrieval performance, namely nearest neighbor (NN), first tier (FT), second tier (ST), F-measure (F),
discounted cumulative gain (DCG), and average normalized modified retrieval rank (ANMRR). 
We refer to ~\citet{benchmark3d} for a more detailed description of these metrics.
Our experiments show, that OSCAR outperforms all other state-of-the-art approaches, despite being the only fully training-free method in this comparison.

\begin{table}[ht]
    \centering
    \setlength{\tabcolsep}{4pt} 
    \caption{Object Retrieval performance on the MI3DOR benchmark. OSCAR is the only fully training-free method compared. Best result in \textbf{bold)}}
    \label{tab:main-results}
    \begin{tabular}{l c c c c c c}
        \toprule
        \textbf{Method} & \textbf{NN $\uparrow$} & \textbf{FT $\uparrow$} & \textbf{ST $\uparrow$} & \textbf{F $\uparrow$} & \textbf{DCG $\uparrow$} & \textbf{ANMRR $\downarrow$} \\
        \midrule
        CORAL~\citep{coral} & 0.362 & 0.174 & 0.256 & 0.060 & 0.199 & 0.816 \\
        MEDA~\citep{MEDA} & 0.430 & 0.344 & 0.501 & 0.046 & 0.361 & 0.646 \\
        JGSA~\citep{jgsa} & 0.612 & 0.443 & 0.599 & 0.116 & 0.473 & 0.541 \\
        \midrule
        JAN~\citep{jan} & 0.446 & 0.343 & 0.495 & 0.085 & 0.364 & 0.647 \\
        RevGrad~\citep{RevGrad} & 0.650 & 0.505 & 0.643 & 0.112 & 0.542 & 0.474 \\
        DLEA~\citep{dlea} & 0.764 & 0.558 & 0.716 & 0.143 & 0.597 & 0.421 \\
        SC-IFA~\citep{sc-ifa} & 0.721 & 0.584 & 0.721 & 0.163 & 0.637 & 0.363 \\
        S2Mix~\citep{s2mix} & 0.841 & 0.670 &0.807 & 0.166 & 0.713 & 0.304 \\
        \midrule
        \textbf{OSCAR (Ours)} & \textbf{0.894} & \textbf{0.708} & \textbf{0.850} & \textbf{0.238} & \textbf{0.844} & \textbf{0.205} \\
        \bottomrule
    \end{tabular}
\end{table}

\subsection{Object Pose Estimation}
\noindent{}We investigate the benefits of OSCAR in the context of novel object pose estimation. Once a vision system is deployed, two options are available. The novel object can be autonomously reconstructed, and the reconstructed model can be used later on for pose estimation. In-the-wild autonomous reconstruction limits the model accuracy. Alternatively, a similar model can be retrieved using OSCAR and substituted for the instance model in the pose estimation process. We propose to evaluate the trade-off between these two strategies.
To validate this, we evaluate the performance of MegaPose~\citep{labbe2022megapose} using three types of 3D object models: 
(i) top-ranked confusion CAD models retrieved by OSCAR from a combined database of YCB-V~\citep{xiang2018posecnn} and GSO~\citep{downs2022google}, (ii) 3D models reconstructed from real-world images, and (iii) ground-truth object models (see Fig.~\ref{fig:retrieved_objects}).
\paragraph{\textbf{Implementation Details}}
\noindent{}To evaluate the next-best-model strategy, we provide MegaPose with the most frequently retrieved \emph{confused model}, i.e., the CAD model that OSCAR incorrectly selects most often for that object class. 
For reconstructed models, we rely on the dataset provided by Burde et al.\citep{burde2025wacv}, which demonstrates that Nerfacto~\citep{mildenhall2021nerf} offers the best trade-off between pose recall and runtime. 
Following their findings, we use the meshes reconstructed with Nerfacto from a set of 50 input images.
As a baseline, we use the ground-truth object models for pose estimation with MegaPose.
All experiments are conducted on a test scene from the YCB-V dataset~\citep{xiang2018posecnn}. Figure~\ref{fig:retrieved_objects} shows the CAD models utilized for each experiment.

\begin{figure}[h]
  \centering
  \includegraphics[width=0.7\columnwidth]{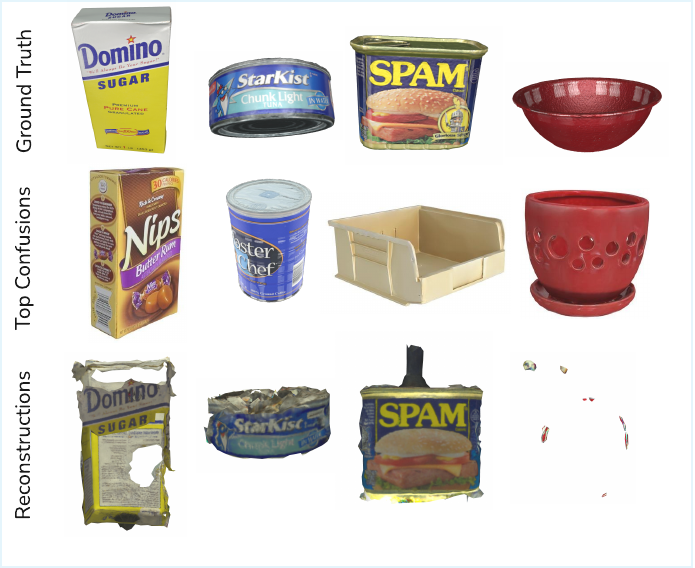}
  \caption{Visualization of object meshes: Ground-truth object models (top), most frequently confused CAD model retrieved by OSCAR (middle), and Nerfacto-reconstructed models (bottom).}
\end{figure}
  \label{fig:retrieved_objects}

\begin{figure*}[th]
    \centering
    \begin{subfigure}[b]{0.23\textwidth}
        \centering
        \hspace{20mm}
        \includegraphics[width=\textwidth, trim=4cm 2.5cm 2.8cm 2.2cm, clip]{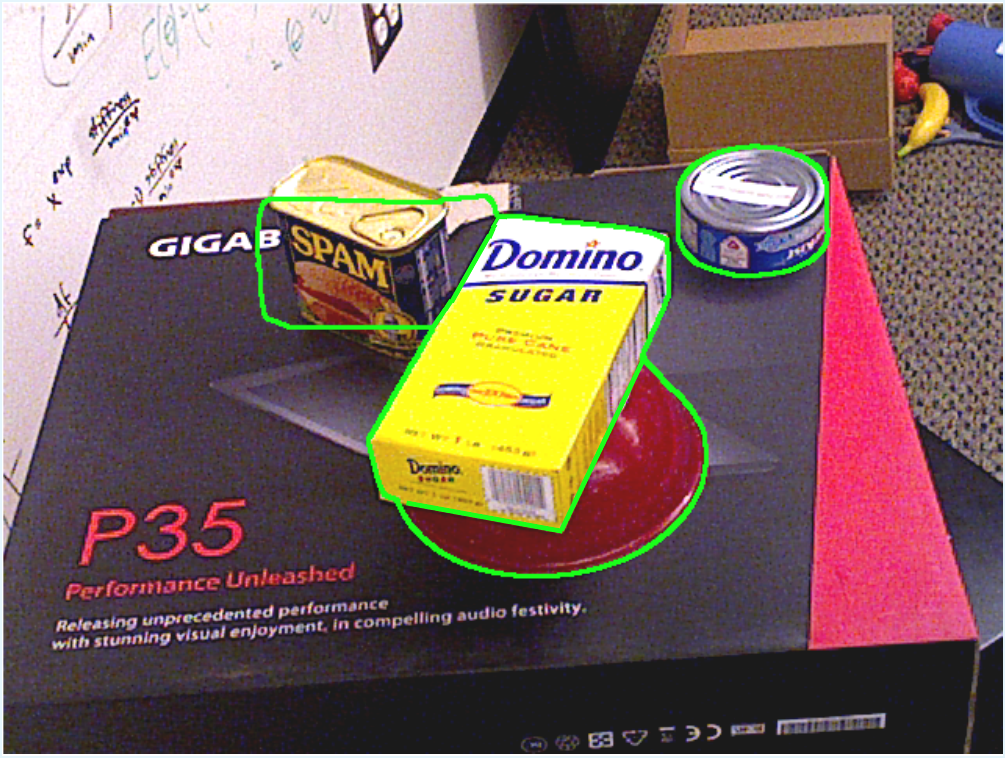}
        \caption{\textbf{Ground-truth models}}
        \label{fig:pose_third}
    \end{subfigure}
    \hfill
    \begin{subfigure}[b]{0.23\textwidth}
        \centering
        \includegraphics[width=\textwidth, trim=4cm 2.5cm 2.8cm 2.2cm, clip]{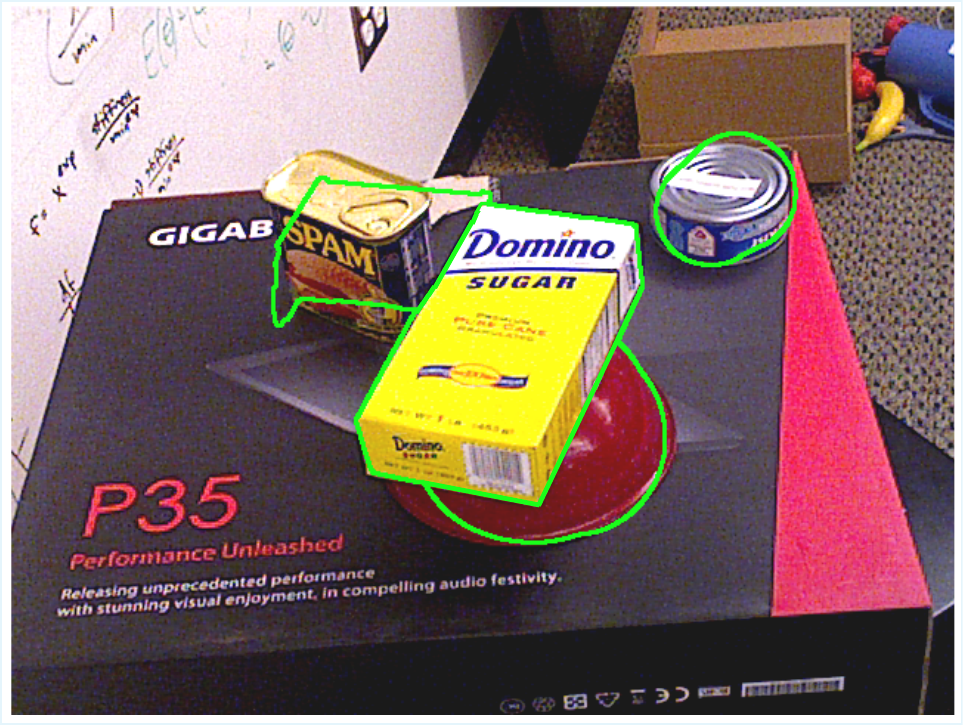}
        \caption{\textbf{CADs retrieved with Oscar}}
        \label{fig:pose_oscar}
    \end{subfigure}
    \hfill
    \begin{subfigure}[b]{0.23\textwidth}
        \centering
        \includegraphics[width=\textwidth, trim=5cm 3cm 3.8cm 3cm, clip]{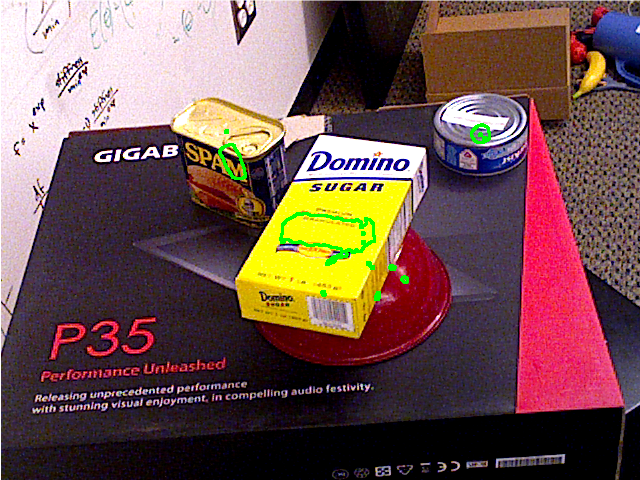}
        \caption{\textbf{Reconstructed CAD models}}
        \label{fig:pose_reconstructed}
    \end{subfigure}\hspace{2mm}
    \caption{Comparison of pose estimation results across three approaches: (a) ground-truth object models, (b) CAD models retrieved by OSCAR, and (c) reconstructed 3D models generated with Nerfacto.}
    \label{fig:pose_estimation}

    \vspace{2mm}
    \small
    \setlength{\tabcolsep}{8pt}
    \renewcommand{\arraystretch}{1.1}
    \begin{tabular}{l|c c c|c c c}
    \toprule
    \textbf{Object} & \multicolumn{3}{c|}{Trl. error \([\mathrm{mm}]\)} & \multicolumn{3}{c}{Rot. error \([\!^\circ]\)} \\
                   & GT & OSCAR & Reconst. & GT & OSCAR & Reconst. \\
            \midrule
            Sugar & 16.29  & 20.19   & 1621.95 & 42.44  & 4.30   & 93.24 \\
            Tuna  & 29.53  & 806.40  & 4760.08 & 22.76  & 51.17  & 94.53 \\
            Bowl  & 25.06  & 85.55   & 2077.02 & 99.83  & 115.23 & 68.59 \\
            Spam  & 185.26 & 665.33  & 3219.92 & 87.83  & 57.17  & 78.39 \\
    \bottomrule
    \end{tabular}
    \caption{Rotation and translation errors for individual objects using OSCAR and reconstructed models against the ground-truth models as baselines. Lower values indicate better performance.}
    \label{tab:pose_errors}
\end{figure*}

\paragraph{\textbf{Results}}
\noindent{}The pose estimation results in Table~\ref{tab:pose_errors} show that CAD models retrieved using OSCAR achieve more accurate pose estimations compared to the reconstruction-based approach. 
Even when using ground-truth models, however, certain objects remain challenging (e.g., the canned meat), highlighting the difficulties of the task. 
Nevertheless, the estimated poses for objects such as the sugar box and the bowl are sufficiently accurate to enable grasping. 
In contrast, when using reconstructed object models, none of the pose estimates proved reliable—MegaPose consistently failed to produce usable results. 
Figure~\ref{fig:pose_estimation} illustrates this comparison, overlaying the estimated poses for CAD models retrieved with OSCAR (Fig.~\ref{fig:pose_oscar}) and for reconstructed meshes (Fig.~\ref{fig:pose_reconstructed}).

\paragraph{\textbf{Runtime Analysis}}
\noindent In our runtime analysis, we evaluate the time required to onboard a novel object instance into the database. 
For comparison with OSCAR, we use a 50-image subset of the image dataset published by~\citep{burde2025wacv}.
We reconstruct the YCB-V Cracker Box using Nerfacto within the Nerfstudio framework on an NVIDIA GeForce RTX 4090 GPU. 
Tab.~\ref{tab:runtime-comparison} outlines the steps involved in generating a 3D mesh, including preprocessing with COLMAP, reconstruction with Nerfacto, and the final mesh extraction from Nerfstudio.
OSCAR onboards every new object by rendering synthetic views of each CAD model and generating textual descriptions.
The time required to add a novel instance to the combined GSO and Housecat6D dataset is also reported in Table~\ref{tab:runtime-comparison}. 
Compared to the reconstruction-based pipeline, OSCAR's onboarding is 31.8x faster, making it significantly more suitable for time-sensitive applications.

\begin{table}[h]
\centering
\small
\captionsetup{width=\columnwidth}
\begin{tabular}{l|c c}
\toprule
\textbf{Stage} & \textbf{Rec.}  & \textbf{OSCAR} \\
\midrule
COLMAP           & 01:07 & --  \\
Reconstruction   & 09:04 & --  \\
Model extraction & 05:43 & -- \\
Text generation  & -- & 0:14  \\
Rendering        & -- & 0:16  \\
\midrule
\textbf{Total}            & \textbf{15:54} & \textbf{00:30} \\
\bottomrule
\end{tabular}
\caption{Comparison of system runtime for onboarding a single novel object to the database [mm:ss]}
\label{tab:runtime-comparison}
\end{table}


\subsection{Ablation Studies: Modalities}
\noindent{}This section presents an evaluation of the impact of the design choices of our retrieval strategy, in particular investigating different retrieval modalities.

\subsubsection{Image-based retrieval}
\noindent{}In the following section, we evaluate the performance of our retrieval framework under different image encoders.
\paragraph{\textbf{Implementation Details}}
\noindent{}We evaluate object retrieval using two different image encoders: CLIP~\citep{radford2021clip}, which leverages joint vision–language pretraining, and DINOv2~\citep{oquab2024dinov2}, a purely vision-based self-supervised model. 
For each object instance, the region of interest (ROI) is extracted either as a bounding box or a segmentation with a gray background to match the rendered images. 
We report mean Average Precision at different $k$ for ($k=1,3,5,10$) to capture both top-1 retrieval accuracy and broader candidate recall. 
\paragraph{\textbf{Results}}
\noindent{}The results in Table~\ref{tab:img2img_map} show that DINOv2 achieves higher top-1 accuracy, particularly on YCB-V (77.38\%) and HCat6D (41.81\%) with segmentation masks, while CLIP with segmentation excels at higher $k$, reaching 92.61\% mAP@10 on YCB-V and performing best on the larger YCB-V\&GSO dataset. 
Segmentation generally improves retrieval, especially for CLIP, though bounding boxes sometimes perform better on HCat6D on similar-looking objects. 
Overall, DINOv2 favors top-1 precision, whereas CLIP with segmentation is more robust for broader candidate retrieval in diverse datasets. 

\begin{table}[h]
\centering
\small
\captionsetup{width=\columnwidth}
\begin{tabular}{l|lccc}
\toprule
 & \textbf{Method} & \textbf{YCB-V} & \textbf{Hcat6D} & \textbf{YCBV\&GSO} \\
\midrule
\multirow{4}{*}{\rotatebox{90}{\scriptsize \textbf{{mAP@1}}}} & CLIP + bbox  & 59.78 & 34.31 & 40.08 \\
                                      & CLIP + segm  & 69.53 & 28.30 & \textbf{41.31} \\
                                      & Dinov2 + bbox  & 70.98 & 33.41 & 31.61\\
                                      & Dinov2 + segm  & \textbf{77.38} & \textbf{41.81} & 36.41 \\
\midrule
\multirow{4}{*}{\rotatebox{90}{\scriptsize \textbf{mAP@3}}} & CLIP + bbox  & 63.82 & \textbf{45.32} & 45.13 \\
                                      & CLIP + segm  & 81.16 & 33.47 & \textbf{47.12}\\
                                      & Dinov2 + bbox  & 75.35 & 39.74 & 36.51 \\
                                      & Dinov2 + segm  & \textbf{81.54} & 44.71 & 40.65 \\
\midrule
\multirow{4}{*}{\rotatebox{90}{\scriptsize \textbf{mAP@5}}} & CLIP + bbox & 63.05 & \textbf{51.11} & 45.29 \\
                                        & CLIP + segm  & \textbf{87.68} & 33.63 & \textbf{46.12}\\
                                      & Dinov2 + bbox  & 74.44 & 39.42 & 36.67\\
                                      & Dinov2 + segm  & 80.23 & 43.10 & 41.33 \\
\midrule
\multirow{4}{*}{\rotatebox{90}{\scriptsize \textbf{mAP@10}}} & CLIP + bbox & 60.03 & \textbf{59.91} & 43.75 \\
                                      & CLIP + segm & \textbf{92.61} & 30.86 & \textbf{44.73} \\
                                      & Dinov2 + bbox & 69.08 & 35.41 & 35.44\\
                                      & Dinov2 + segm  & 75.20 & 39.21 & 40.37 \\
\bottomrule
\end{tabular}
\caption{Comparison of Top-$k$ mean Average Precision for image-to-image across YCB-V, Housecat6D, and the combined YCB-V\&GSO datasets. Best results for each $k$ and each dataset are printed \textbf{bold}.}
\label{tab:img2img_map}
\end{table}
\subsubsection{Text-based Retrieval}
\begin{figure}[tbp]
\centering

\includegraphics[width=0.2\textwidth]{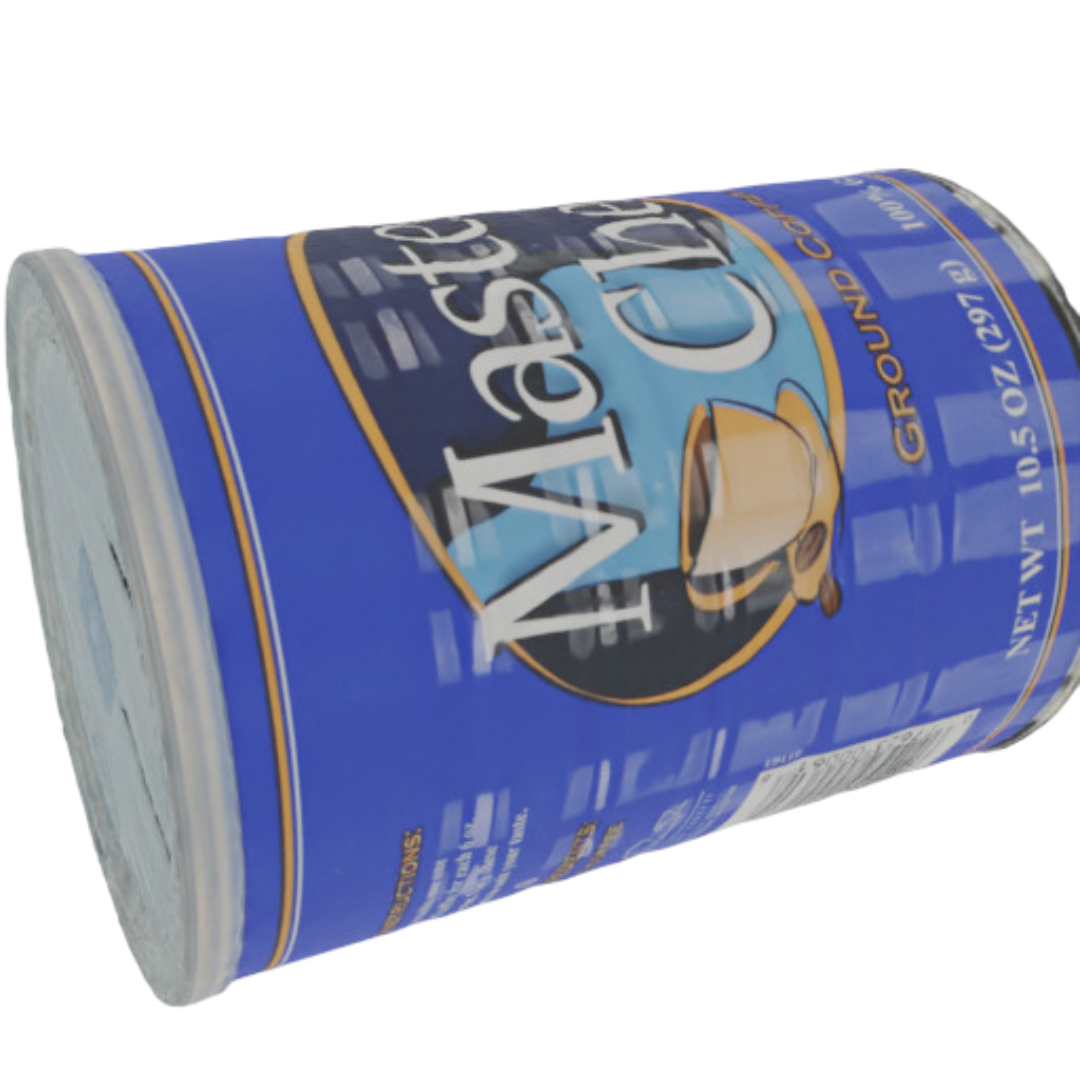}

\renewcommand{\arraystretch}{1.2}
\rowcolors{2}{gray!10}{white}
\footnotesize
\resizebox{1.1\textwidth}{!}{
\begin{tabularx}{\textwidth}{p{1.2cm} X X} 
    \toprule
    \textbf{Type} & \textbf{Prompt} & \textbf{Example Output} \\
    \midrule
    \textbf{blind} & Imagine you're describing the object to a blind person. Be extremely detailed about the object's appearance, colors, shape, material, any text or logos, and any unique markings. & The object is a blue can with a black top. The can is cylindrical in shape and has a black top that is slightly curved. The can is made of metal, and it is filled with a substance, possibly coffee. The can has a logo on it, which is a picture of a coffee cup. The can is also labeled with the word "Major" and the number "100." The overall appearance of the can is that of a standard coffee. \\
    \textbf{comma} & Describe the object in a comma-separated list, focusing on its visual appearance, color, material, and any visible text or brand names. Be concise. & Blue, silver, black, and white can. \\
    \textbf{caption} & Write a detailed visual caption of the image, mentioning colors, materials, brand names, and visible labels. & The image features a large blue can of coffee, possibly a can of Maxwell House coffee. The can is prominently displayed in the foreground, occupying a significant portion of the image. The can is made of metal and has a label on it, indicating the brand and the contents inside. The overall scene is a close-up of the coffee can, showcasing its size and design. \\
    \textbf{attributes} & Extract visual attributes of the main object in the image: object type, brand name, color, material, and label text. & The main object in the image is a blue can with a label. The can is made of metal and has a black label. The brand name is "Major." The can is filled with a product, possibly coffee, as indicated by the label text. \\
    \bottomrule
\end{tabularx}
}

\caption{Example image (top) and table (bottom) comparing four prompting strategies. The table lists the prompt type, prompt text, and an example output.}
\label{tab:prompt_examples}
\end{figure}
\noindent{}For the text-based retrieval strategy, we consider four strategies describing the objects with different levels of detail.
For each rendered image, we generate a description with LLaVa-v1.5-7B~\citep{liu2023improvedllava}. 
Table~\ref{tab:prompt_examples} shows the used prompts with example outputs.
All methods receive ground-truth bounding boxes as input and operate under the assumption of access to a 3D object database. 
Table~\ref{tab:language_features} shows that prompt choice strongly affects retrieval performance. 
The blind prompt performs best on a distinct dataset such as YCB-V, while the attributes prompt achieves higher scores on the larger datasets Housecat6D and YCB-V\&GSO. 
Caption- and comma-style prompts generally underperform, indicating that detailed or structured descriptions provide more discriminative cues for text-based 3D object retrieval.

\begin{table}[!ht]
\centering
\small
\captionsetup{width=\columnwidth}
\begin{tabular}{l|lccc}
\toprule
 & \textbf{Method} & \textbf{YCB-V} & \textbf{Hcat6D} & \textbf{YCBV\&GSO} \\
\midrule
\multirow{4}{*}{\rotatebox{90}{\scriptsize \textbf{mAP@1}}} & blind & \textbf{68.12} & 16.87 & 30.33 \\
                                      & comma & 15.88 & 4.97 & 6.42 \\
                                      & caption & 43.52 & 16.11 & 31.39 \\
                                      & attributes & 32.95 & \textbf{25.89} & \textbf{46.11} \\
\midrule
\multirow{4}{*}{\rotatebox{90}{\scriptsize \textbf{mAP@3}}} & blind & \textbf{72.75} & 20.82 & 34.91 \\
                                      & comma & 18.04 & 5.87 & 7.18 \\
                                      & caption & 46.38 & 19.71 & 34.11 \\
                                      & attributes & 34.88 & \textbf{28.98} & \textbf{50.51} \\
\midrule
\multirow{4}{*}{\rotatebox{90}{\scriptsize \textbf{mAP@5}}} & blind & \textbf{72.50} & 21.20 & 34.55 \\
                                      & comma & 15.57 & 5.89 & 7.25 \\
                                      & caption & 46.38 & 20.25 & 33.80 \\
                                      & attributes & 35.04 & \textbf{29.37} & \textbf{50.40} \\
\midrule
\multirow{4}{*}{\rotatebox{90}{\scriptsize \textbf{mAP@10}}} & blind & \textbf{69.63} & 20.43 & 32.71 \\
                                      & comma    & 18.90 & 5.80 & 7.10 \\
                                      & caption    & 44.50 & 19.57 & 32.20 \\
                                      & attributes   & 33.99 & \textbf{28.40} & \textbf{48.58} \\
\bottomrule
\end{tabular}
\caption{Comparison of mAP-$k$ for different language prompts across YCB-V, Housecat6D, and the combined YCB-V\&GSO datasets. Best results for each $k$ and each dataset are printed \textbf{bold}.}
\label{tab:language_features}
\end{table}

\subsection{Ablation Studies: Finetuning}
\noindent{}Based on the results from both image-based and text-based retrieval, we observed that each modality has notable limitations. 
To leverage the strengths of both, we first filter candidate instances by matching textual descriptions to the input image, and then identify the best-matching reference image using DINOv2. 
Figures \ref{fig:accuracy_threshold} and \ref{fig:topk_accuracy_lines} present the results of our ablation studies focusing on the impact of threshold selection and top-k filtering on model accuracy across different datasets.

\subsubsection{Thresholding}
\begin{figure}[h]
    \centering
    \begin{tikzpicture}
    \begin{axis}[
        width=0.65\columnwidth,  
        height=0.4\columnwidth, 
        xlabel={Threshold},
        ylabel={Average Precision (\%)},
        grid=both,
        grid style={line width=.1pt, draw=gray!20},
        major grid style={line width=.2pt,draw=gray!50},
        ymin=0, ymax=100,
        xmin=0.09, xmax=0.7,
        legend style={at={(0.5, 0.25)}, anchor=north, legend columns=-1}
    ]

    \addplot[
        color=blue!70,
        mark=o,
        thick
    ] coordinates {
        (0.10,36.41)
        (0.15,36.41)
        (0.20,36.41)
        (0.25,36.22)
        (0.30,37.50)
        (0.33,41.09)
        (0.35,54.50)
        (0.36,57.72)
        (0.37,60.00)
        (0.38,58.91)
        (0.39,58.45)
        (0.40,54.50)
        (0.45,43.27)
        (0.50,42.98)
        (0.55,42.98)
        (0.60,42.98)
        (0.65,42.98)
    };
    \addlegendentry{YCBV\&GSO}

    \addplot[
        color=red!70,
        mark=o,
        thick
    ] coordinates {
        (0.10,77.38)
        (0.15,77.38)
        (0.20,77.38)
        (0.25,77.60)
        (0.30,80.16)
        (0.35,84.75)
        (0.37,86.23)
        (0.38,86.72)
        (0.39,86.62)
        (0.40,85.93)
        (0.45,81.92)
        (0.50,81.60)
        (0.55,81.60)
        (0.60,81.60)
        (0.65,81.60)
    };
    \addlegendentry{YCBV}

    \addplot[
        color=green!60!black,
        mark=o,
        thick
    ] coordinates {
        (0.10,41.46)
        (0.15,41.46)
        (0.20,41.47)
        (0.25,41.64)
        (0.30,45.32)
        (0.35,48.09)
        (0.36,48.67)
        (0.37,48.63)
        (0.38,47.94)
        (0.39,48.05)
        (0.40,48.33)
        (0.45,48.30)
        (0.50,48.30)
        (0.55,48.30)
        (0.60,48.30)
        (0.65,48.30)
    };
    \addlegendentry{HCAT}

    \end{axis}
    \end{tikzpicture}
    \caption{Average Precision vs. Threshold for Filtering candidate objects for YCB-V, YCB-V\&GSO and Housecat6D datasets}
    \label{fig:accuracy_threshold}
\end{figure}
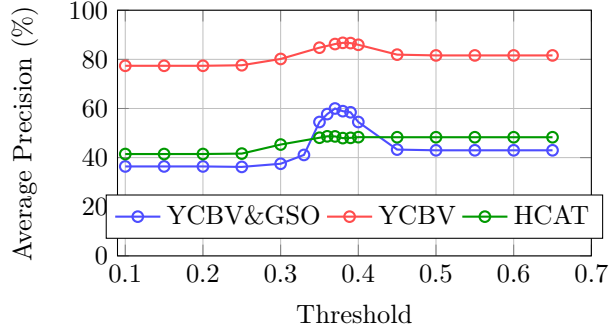

\noindent{}The threshold defines at which clip-based cosine similarity a corresponding object is selected as candidate.
Figure~\ref{fig:accuracy_threshold} illustrates the effect of varying the threshold parameter on the accuracy for three datasets: YCBV\&GSO, YCBV, and Housecat6D.
We observe that for YCBV\&GSO, accuracy remains relatively stable at lower thresholds (0.10–0.30), followed by an increase around thresholds 0.35–0.37, reaching a peak at 60\%, before declining at higher thresholds. A similar trend is evident for YCBV, with accuracy peaking at 86.72\% at a threshold of 0.38, while Hcat6D demonstrates a more gradual improvement, plateauing around 48\%.
These trends suggest that the performance can be maximized with a threshold of 0.37.

\subsubsection{Tok-k Filtering}
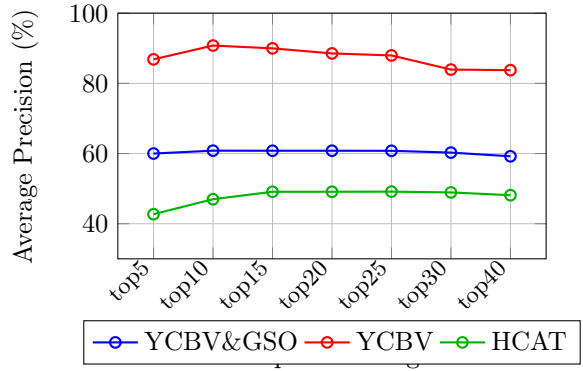
\begin{figure}[h]
    \centering
    \begin{tikzpicture}
    \begin{axis}[
        width=0.6\columnwidth,
        height=0.4\columnwidth,
        xlabel={Top-k Filtering},
        ylabel={Average Precision (\%)},
        ymin=30, ymax=100,
        xtick={1,2,3,4,5,6,7},
        xticklabels={top5, top10, top15, top20, top25, top30, top40},
        xlabel style={at={(0.5,-0.15)}, anchor=north},
        xticklabel style={rotate=45, anchor=east, font=\small},
        grid=both,
        grid style={line width=.1pt, draw=gray!20},
        major grid style={line width=.2pt,draw=gray!50},
        legend style={at={(0.5,-0.25)}, anchor=north, legend columns=-1}
    ]

    \addplot[
        color=blue,
        mark=o,
        thick
    ] coordinates {
        (1,60)
        (2,60.82)
        (3,60.8)
        (4,60.8)
        (5,60.78)
        (6,60.27)
        (7,59.22)
    };
    \addlegendentry{YCBV\&GSO}
    
    \addplot[
        color=red,
        mark=o,
        thick
    ] coordinates {
        (1,86.84)
        (2,90.78)
        (3,89.97)
        (4,88.53)
        (5,87.96)
        (6,83.93)
        (7,83.76)
    };
    \addlegendentry{YCBV}

    \addplot[
        color=green!70!black,
        mark=o,
        thick
    ] coordinates {
        (1,42.72)
        (2,46.99)
        (3,49.11) 
        (4,49.11)
        (5,49.15)
        (6,48.92) 
        (7,48.13)
    };
    \addlegendentry{HCAT}

    \end{axis}
    \end{tikzpicture}
    \caption{Average Precision vs Top-k Filtering for YCB-V, YCB-V\&GSO and Housecat6D datasets}
    \label{fig:topk_accuracy_lines}
\end{figure}

\noindent{}If none of the retrieved objects surpasses the confidence threshold, we employ a top-k filtering strategy to select candidate models for evaluation.
Figure~\ref{fig:topk_accuracy_lines} illustrates how the average precision varies as the number of top-k candidates increases for the three datasets: YCB-V, Housecat6D, and YCB-V\&GSO.
Overall, performance remains relatively stable across most values of k, indicating that the retrieval method is robust to moderate changes in the number of candidates considered.
For YCB-V, accuracy peaks around the top-10 candidates (90.8\%) and gradually decreases for larger k, suggesting that additional retrieved models tend to introduce more noise than useful matches.
A similar pattern is observed for YCB-V\&GSO and Housecat6D, where performance slightly improves up to top-10 and then declines marginally for larger k.

\section{Conclusion}

\noindent{}We presented OSCAR, a novel method that retrieves a matching object model from an unlabeled 3D object database using a single RGB image and a natural language prompt.
We demonstrated that OSCAR retrieves object models reliably by combining text embeddings and image embeddings in a two-stage process.
Furthermore, we proposed a novel pose estimation strategy alternative to time-costly reconstruction-based approaches.
Integrating OSCAR with existing pose estimation techniques yields improved accuracy and efficiency, demonstrating the effectiveness of retrieval paradigms in practical scenarios where the exact CAD model or multi-view data are unavailable. 
Future work will aim to embed OSCAR in a pose estimation framework.

\section*{Author contributions: CRediT}
\noindent{}\textbf{Tessa Pulli}: Conceptualization, Investigation, Writing-original-draft. \textbf{Jean-Baptiste Weibel}: Writing-review-editing. \textbf{Peter Hönig}: Writing-review-editing.\textbf{Matthias Hirschmanner}: Writing-review-editing. \textbf{Markus Vincze}: Writing-review-editing, Supervision, Project-administration, Funding-acquisition. \textbf{Andreas Holzinger}: Writing-review-editing.

\section*{Acknowledgement}

\noindent{}We gratefully acknowledge the support of the EU-program EC Horizon 2020 for Research and
Innovation under project No. I 6114, project iChores.

\section*{Declaration of generative AI and AI-assisted technologies in the manuscript preparation process}
\noindent{}During the preparation of this work the authors used ChatGPT and Google Gemini in order to improve the language and readability of the manuscript, and to assist in writing Python code for visualizing experimental results.
After using these tools, the authors reviewed and edited the content as needed and take full responsibility for the content of the published article.

\bibliographystyle{elsarticle-harv} 
\bibliography{bib}



\end{document}